# Decision Making with Side Information and Unbounded Loss Functions

Majid Fozunbal and Ton Kalker
Hewlett-Packard Laboratories
Pale Alto, CA 94304

January 2006

**Abstract**

We consider the problem of decision-making with *side information* and *unbounded loss functions*. Inspired by probably approximately correct learning model, we use a slightly different model that incorporates the notion of side information in a more generic form to make it applicable to a broader class of applications including parameter estimation and system identification. We address sufficient conditions for consistent decision-making with exponential convergence behavior. In this regard, besides a certain condition on the growth function of the class of loss functions, it suffices that the class of loss functions be dominated by a measurable function whose *exponential Orlicz expectation* is uniformly bounded over the probabilistic model. Decay exponent, decay constant, and sample complexity are discussed. Example applications to method of moments, maximum likelihood estimation, and system identification are illustrated, as well.

## 1 Introduction

Decision-making refers to a system-theoretic problem of choosing among alternative options in light of their possible outcomes. Analytically, it is an optimization problem where the decision-maker's goal is to find optimal or suboptimal points, known as *actions*, to attain objectives that are quantified by certain utility functions. Most often, depending on the complexity of an underlying system, in particular in stochastic systems, a decision-maker suffers from uncertainties in determining the outcome of an action. In such systems, a decision-maker can use side information to reduce the risk in making decisions. This basically is a common scenario in applications such as learning theory, hypothesis testing, inventory control, parameter estimation, pattern recognition, system identification, prediction, and filtering. Hence, analytical results and arguments within the generic model of decision-making under uncertainty provide important insights and tools in treating the aforementioned applications.

In this regard, learning theory has been one of the major areas that has been treated from a decision-making viewpoint [1]. Learnability or *consistent decision-making* is a property that is attributed to a *class of loss functions* that satisfy certain analytical requirements. In learning theory, whether supervised or unsupervised, the true risk value for an action, known as generalization error, is measured as the expected *loss* over the observation space, where the governing probability distribution belongs to a known class of probability measures called the *probabilistic model*. One of the main challenges in verification of learnability is to prove the existence of a *learning algorithm* or *decision policy* that can use a sequence of observations and take actions whose true risk converge to the minimal risk as the number of observations grows to infinity. In this regard, the original model for learnability introduced by Valiant [2], known as *probably approximately correct* (PAC) learning, and its decision-making based generalization by Haussler [1] have been shown to be strongly related to the existence of uniform laws of large numbers [3], [4].

Much work in this area addresses and discusses consistency for the case of *bounded* loss functions. The importance of bounded loss functions is that they allow consistency analysis over *universal probabilistic models*, also known as *distribution free models*. To this end, the most commonly used decision policy is the *empirical risk minimization* (ERM) method [4]. It is known that the ERM policy is universally consistent if the collection of the *level sets* of the class of loss functions has a finite *VC-dimension* [4], [5]. While these results on bounded loss functions provide important insights on applications such as pattern recognition and neural networks, many other applications including regression analysis, parameter estimation, and system identification motivate consistency analysis of decision-making with unbounded loss functions. Although there are some sufficient conditions for *relative* uniform convergence of empirical risk functions [4, Ch. 5], results on (*absolute*) uniform convergence are more insightful and desirable in assessing the consistency of the ERM policy.

With this motivation, and inspired by [1], in this work we study the problem of consistent decision-making using unbounded loss functions. To state and formulate the problem, we use a model that is slightly different from the one commonly used in learning theory. This model explicitly uses generic notions of *solution defining operator* and the *side information*. Adopting such an approach, we aim not only to make the final results span a larger extent of applications, but also to provide a more intuitive method in articulating other applications into this model.

We provide sufficient conditions for consistency of the ERM policy with exponential convergence behavior. These conditions are also sufficient for *strong consistency* of the ERM policy. Unlike the case of bounded loss functions, in this case, besides the behavior of the growth function of the class of loss functions, asymptotic behaviors of the probabilities of the level sets of the loss functions are important in determining consistency and convergence behavior of the ERM policy. More precisely, it is shown that to have an exponential convergence behavior, including a certain condition on the growth rate of the class of loss functions, similar to the case of bounded loss functions, it suffices to have a *dominating function* (for the class of loss functions) whose *exponential Orlicz expectation* is uniformly bounded over the probabilistic model. In contrast to the case of bounded loss functions, the *decay exponent* is strictly smaller than one. On the other hand, the expression of *decay constant* is very similar to its counterpart in the bounded case. *Sample complexity* and the effect of different parameters on it are also addressed. Furthermore, example applications to method of moments, maximum likelihood estimation, and system identification are illustrated.

## 2 Analytical Setup

Let $\mathcal{H}$ be a subset of a separable normed space that denotes the *state space* of a given system. For example, in channel estimation for a mobile communication system, $\mathcal{H}$ is defined as the inclusion of all possible channel transfer functions between the transmitter and receiver. In parameter estimation, $\mathcal{H}$ is defined as the inclusion of all unknown parameters of interest. Although in the provided examples, one seeks to find a good estimate for the unknown channel state or the unknown parameter to minimize some error criteria, in general, one seeks a good *action* to attain certain objectives. Let $\mathcal{G}$ be a compact subset of a Euclidean space called the *action space*, where one takes an action $g \in \mathcal{G}$ to attain an objective, measured as expected loss.

In stochastic systems, associated with any state and any action, there may exist a range of possible outcomes whose likelihood is determined by the probabilistic model that governs all uncertain underlying parameters affecting the outcome. For example, in a linear additive noise channel, the additive noise is the uncertain parameter that affects the output of the channel, i.e., the outcome. Let $\mathcal{U}$ be a subset of a separable normed space that models the range of such parameters. We call $\mathcal{U}$ the *uncertainty space*.

Suppose the system is in state $h \in \mathcal{H}$ and the uncertain parameter is $u \in \mathcal{U}$. If there exists an oracle who provides perfect information about $h$ and $u$, one could uniquely determine the outcome for each $g \in \mathcal{G}$ and pick the action that is most favorable. In practice, however, elements $h$ and $u$ are not given and instead some *side information* about them is known. Let $\mathcal{O}$ be a subset of a Euclidean space called the *observation space*. Side information is generated by a sequence of mappings $\boldsymbol{I} = (I_n)_{n \in \mathbb{N}}$ such that for each $n$, $I_n \colon \mathcal{H} \times \mathcal{U}^n \to \mathcal{O}^n$, where $I_n$ is formed as a stationary extension of a function $I \colon \mathcal{H} \times \mathcal{U} \to \mathcal{O}$ called the *information function*. For an underlying pair $(h, u^n)$, $I_n$ generates a multi-sample $o^n = (o_i)_{i=1}^n \in \mathcal{O}^n$ as side information. Generation of the side information is assumed to be governed by an independent, identically distributed (i.i.d.) stochastic process as follows.

Let $(\Omega, \mathscr{F})$ be a *measurable space*, where $\Omega$ stands for a *sample space* and $\mathscr{F}$ denotes its Borel $\sigma$-field. Let $\mathscr{P}_\Theta = \{P_\theta : \theta \in \Theta\}$ be a collection of probability measures defined over $(\Omega, \mathscr{F})$. For each $\theta$, let $\{U_n^\theta\}_{n \in \mathbb{N}}$ denote an i.i.d. stochastic process ranged over $\mathcal{U}$ whose distribution is induced from $P_\theta$. As a result of information function, for each $h \in \mathcal{H}$, we obtain an i.i.d. stochastic process $\{O_n^{\theta,h}\}$, called the *side information process*, where at each time instance $n$, $O_n^{\theta,n}\colon \Omega \to \mathcal{O}$ is a random element whose induced probability measure is denoted by $P_{\theta,h}$. For every Borel-measurable $E \subseteq \mathcal{O}$, $P_{\theta,h}(E)$ is described by:

$$P_{\theta,h}(E) = P_\theta(u \in \mathcal{U} : (h, u) \in I^{-1}(E)). \tag{1}$$

We denote the class of all induced probability measures on $\mathcal{O}$ by $\mathscr{P}_{\Theta,\mathcal{H}}$. We assume that for every bounded subset $E \subset \mathcal{O}$, there exists a probability measure $P_{\theta,h} \in \mathscr{P}_{\Theta,\mathcal{H}}$ such that $P_{\theta,h}(E) < 1$. This condition ensures that the collection $\mathscr{P}_{\Theta,\mathcal{H}}$ is not of bounded support.

Let $Q\colon \mathcal{O} \times \mathcal{G} \to \mathbb{R}$ be a lower bounded, lower semicontinuous *loss function* that determines the penalty of taking action $g \in \mathcal{G}$ when the side information $o \in \mathcal{O}$ is given. We denote the *class of loss functions* by $\mathscr{Q} = \{Q(\cdot, g)\}_{g \in \mathcal{G}}$. We assume all the elements of $\mathscr{Q}$ are measurable and for every $g \in \mathcal{G}$, there exists no distinct $g' \in \mathcal{G}$ such that $Q(o, g') \leq Q(o, g)$ for all $o \in \mathcal{O}$.

Suppose the underlying probability measure is $P_\theta$ and the system is in state $h$. As the generation of side information is governed by $P_{\theta,h}$, the *risk* (expected loss) of an action $g \in \mathcal{G}$ is determined by

$$J_\theta(h, g) = \int Q(o, g) dP_{\theta,h}. \tag{2}$$

The objective of a decision-maker is to minimize the risk in decision-making. An action with lower risk is more favorable and an action that minimizes the risk (upon existence) is an optimal action. Depending on the application, a decision-maker is also interested in actions whose risks are close to the minimal risk to some desired accuracy. Thus, it is of interest to investigate whether a decision-maker can take such actions with a desired confidence in obtained risk.

**Definition 2.1 (Solution defining operator).** *Let $J_\theta(h, \cdot)$ be lower semicontinuous over $\mathcal{G}$ for every $\theta \in \Theta$ and $h \in \mathcal{H}$. For every $h \in \mathcal{H}$, $\theta \in \Theta$, and accuracy $\varepsilon \geq 0$, let*

$$\mathrm{S}_\theta(h, \varepsilon) = \Big\{g \in \mathcal{G} : J_\theta(h, g) \leq \inf_{g' \in \mathcal{G}} J_\theta(h, g') + \varepsilon\Big\} \tag{3}$$

*denote the $\varepsilon$-approximate $\theta$-solution set for $h$. We call $\mathrm{S}\colon \mathcal{H} \times \Theta \times \mathbb{R}^+ \to 2^\mathcal{G}$ a solution defining operator.*

To illustrate how Definition 2.1 can be used in practice, consider the following examples.

**Example 2.1 (Minimum mean square error).** *Consider a linear system with input-output relation*

$$y = \langle h, x \rangle + v,$$

*where $h \in \mathcal{H} \subset \mathbb{R}^m$, $x \in \mathcal{X} = \mathbb{R}^m$, $y \in \mathcal{Y} = \mathbb{R}$, $v \in \mathcal{V} = \mathbb{R}$ denote its state vector, input vector, output value, and noise value, respectively. Let $\mathcal{H}$ be compact and let $\mathcal{G} = \mathcal{H}$. Suppose $\mathscr{P}_\Theta$ is a tight collection of probability measures. Taking*

$\mathcal{U} = \mathcal{X} \times \mathcal{V}$, $\mathcal{O} = \mathcal{X} \times \mathcal{Y}$, and $Q((x,y), g) = |y - \langle g, x \rangle|^2$, *the* mean square error $\varepsilon$-approximate $\theta$-solution set for h is

$$S_\theta(h, \varepsilon) = \left\{ g \in \mathcal{G} : \int |y - \langle g, x \rangle|^2 \, dP_{\theta, h} \leq \inf_{g' \in \mathcal{G}} \int |y - \langle g', x \rangle|^2 \, dP_{\theta, h} + \varepsilon \right\}. \quad (4)$$

**Example 2.2 (Maximum likelihood).** *Let $(\Omega, \mathscr{F}, P)$ be the underlying probability space, i.e., $\Theta$ is a singleton. Let $\mathcal{X} \subseteq \mathbb{R}^m$ and $\mathcal{H}$ be a compact subset of a Euclidean space. For every $h \in \mathcal{H}$, let $X_h$ be random variable over $\mathcal{X}$ whose induced probability measure $P_h$ is absolutely continuous with a continuous density function $f_h(x)$ such that $\sup_{h,x} f_h(x) < \infty$. Taking $\mathcal{G} = \mathcal{H}$, $\mathcal{U} = \Omega$, $\mathcal{O} = \mathcal{X}$, and $Q(x, g) = -\log_2 f_g(x)$, the* maximum likelihood *$\varepsilon$-approximate solution set for h is defined*

$$S(h, \varepsilon) = \left\{ g \in \mathcal{G} : -\int \log_2 f_g dP_h \leq \inf_{g' \in \mathcal{G}} -\int \log_2 f_{g'} dP_h + \varepsilon \right\}. \quad (5)$$

As it is evident in Definition 2.1 as well as the aforementioned examples, determination of $\varepsilon$-approximate $\theta$-solution sets are dependent on the knowledge of the underlying probability measure, $P_{\theta, h}$. If an oracle reveals $P_{\theta, h}$, for any accuracy $\varepsilon \geq 0$, a decision-maker is able to find an action that belongs to $S_\theta(h, \varepsilon)$. In practice, however, such an oracle does not exist. Instead, we are provided with some side information in the form of a sequence of observations $o^n = (o_1, o_2, \ldots, o_n)$, and we seek an action $g \in \mathcal{G}$ whose risk, with high probability, is within $\varepsilon$ accuracy of the minimum risk for sufficiently large $n$. In other words, for each $n$, it is of interest to take an action $g_n$ such that all but a finite number of elements of the sequence $(g_n)_{n \in \mathbb{N}}$ lie, with arbitrary high probability, in $S_\theta(h, \varepsilon)$. This requirement is called *consistency* that is more precisely expressed as follows.

**Definition 2.2 (Consistent decision policy).** *Let $\boldsymbol{A} = (A_n)$ be a sequence of decision rules $A_n \colon \mathcal{O}^n \to \mathcal{G}$ such that $A_n(o^n)$ is an element of $\mathcal{G}$. We call $\boldsymbol{A}$ a* uniformly $\varepsilon$-consistent decision policy *if there exists $\varepsilon$ such that*

$$\limsup_{n \to \infty} p(\boldsymbol{A}, \varepsilon, n) = 0, \quad (6)$$

*where*

$$p(\boldsymbol{A}, \varepsilon, n) \triangleq \sup_{\theta \in \Theta} \sup_{h \in \mathcal{H}} P_\theta(\omega \in \Omega \colon A_n(O^n(\omega)) \notin S_\theta(h, \varepsilon)) \quad (7)$$

*is the worst case probability that the risk associated with decision rule $A_n$ is not within $\varepsilon$ accuracy of the minimal risk. Correspondingly, $\boldsymbol{A}$ is said to be* uniformly consistent *if it is uniformly $\varepsilon$-consistent for every $\varepsilon > 0$. Moreover, if instead of* convergence in probability, almost sure convergence *occurs, the decision policy is said to be* strongly consistent.

Definition 2.2 provides important tools to assess the performance of a decision policy. For example, for a system identification algorithm, convergence in the sense of (6) implies that for every desired level of accuracy, there exists $n_o$ such that if the length of side information is $n \geq n_o$, the algorithm attains an estimate of the system's state within the desired accuracy with arbitrarily high probability. Intuitively, the smaller the $n_o$ is, the more efficient the algorithm is using the side information.

## 3   Empirical Risk Minimization

Any sequence of decision rules that satisfies the conditions of Definition 2.2 is a consistent decision policy. However, verification of its consistency and investigation of its convergence behavior may not be theoretically possible. As a result, much of the work in literature, in particular in learning theory, has been concentrated on a certain form of decision policy that is based on empirical risk minimization (ERM) denoted by $\boldsymbol{A}_{\mathrm{ERM}}$ [4].

Let $o^n = O^n(\omega)$ denote the beginning segment of a realization of the side information process. The *empirical risk function* is defined as

$$J^{(n)}(g) = \frac{1}{n} \sum_{i=1}^{n} Q(g, o_i). \tag{8}$$

Let $\nu^{(n)} = \inf_{g \in \mathcal{G}} J^{(n)}(g)$. Decision rule $A_n$ takes an action

$$\mathrm{A}_n(o^n) \in \{g \in \mathcal{G} : J^{(n)}(g) = \nu^{(n)}\}, \tag{9}$$

as the image of side information $o^n$ in the action space. We now state and prove a result that indicates that to prove $\varepsilon$-consistency of the ERM decision policy, it suffices to prove the uniform convergence of the empirical risk functions.

**Lemma 3.1.** *For every $\theta \in \Theta$, $h \in \mathcal{H}$, and $\varepsilon \geq 0$, the following inequality holds true*

$$P_\theta(\omega \in \Omega : A_n(O^n(\omega)) \notin \mathrm{S}_\theta(h, \varepsilon))$$
$$\leq P_\theta(\omega \in \Omega : \sup_{g \in \mathcal{G}} |J_\theta(h, g) - J^{(n)}(g)| \geq \varepsilon/2). \tag{10}$$

*Proof.* Let $J_\theta^o(h) = \inf_{g \in \mathcal{G}} J_\theta(h, g)$. One can verify that

$$\Psi_{n,\theta}(h, \varepsilon, \delta) \triangleq \{\omega \in \Omega : A_n(O^n(\omega)) \notin \mathrm{S}_\theta(h, \varepsilon)\}$$
$$= \{\omega \in \Omega : \exists g \in \mathcal{G}, J^{(n)}(g) = \nu^{(n)}, J_\theta(h, g) > J_\theta^o(h) + \varepsilon\}.$$

Let define

$$\Delta_n(\varepsilon) \triangleq \left\{\omega \in \Omega : J_\theta^o(h) - \nu^{(n)} > -\varepsilon/2\right\}$$

and its complement $\Delta_n^c(\epsilon) = \Omega \backslash \Delta_n(\epsilon)$. One can verify that

$$\Psi_{n,\theta}(h,\varepsilon,\delta) \subseteq (\Psi_{n,\theta}(h,\varepsilon,\delta) \cap \Delta_n(\epsilon)) \cup \Delta_n^c(\epsilon). \tag{11}$$

Consider the event $\Psi_{n,\theta}(h,\varepsilon,\delta) \cap \Delta_n(\epsilon)$ and let $\omega$ be an outcome belonging to this event. By definition, there exists a $g \in \mathcal{G}$ that satisfies the conditions of $\Psi_{n,\theta}(h,\varepsilon,\delta)$. We can verify that for outcome $\omega$, the inequality

$$J_\theta(h,g) - J^{(n)}(g) > \varepsilon + J_\theta^o(h) - \nu^{(n)} > \varepsilon/2$$

holds true. As a result, for every outcome $\omega \in \Psi_{n,\theta}(h,\varepsilon,\delta) \cap \Delta_n(\epsilon)$, we have

$$\sup_{g \in \mathcal{G}} \big(J_\theta(h,g) - J^{(n)}(g)\big) > \varepsilon/2, \tag{12}$$

which implies that

$$\Psi_{n,\theta}(h,\varepsilon,\delta) \cap \Delta_n(\epsilon) \subseteq \{\omega \in \Omega : \sup_{g \in \mathcal{G}} \big(J_\theta(h,g) - J^{(n)}(g)\big) \geq \varepsilon/2\}. \tag{13}$$

Now consider an outcome $\omega \in \Delta_n^c(\varepsilon)$ and note that

$$v^{(n)} - J_\theta^o(h) \geq \varepsilon/2 \tag{14}$$

holds for this outcome. Suppose $J_\theta(h,\cdot)$ attains its minimum at a point $g^o \in \mathcal{G}$ with the minimum value of $J_\theta^o(h) = J_\theta(h,g^o)$. Noting that $J^{(n)}(g^o) \geq \nu^{(n)}$, we have $J^{(n)}(g^o) - J_\theta(h,g^o) \geq \varepsilon/2$ that implies $\sup_{g \in \mathcal{G}} \big(J^{(n)}(g^o) - J_\theta(h,g^o)\big) \geq \varepsilon/2$. As a result, we deduce that

$$\Delta_n^c(\varepsilon) \subseteq \{\omega \in \Omega : \sup_{g \in \mathcal{G}} \big(J^{(n)}(g) - J_\theta(h,g)\big) \geq \varepsilon/2\}. \tag{15}$$

Using (11) along with (13) and (15), we can conclude the assertion. Note that the right hand sides (RHS) of both (13) and (15) are measurable sets as we assumed both $\mathcal{G}$ and $\mathscr{Q}$ are separable. Hence, in asserted inequality, the outer measure in the left hand side (LHS) is replaced by measure in the RHS. □

To investigate the consistency of the ERM policy, it suffices to show that the right hand side (RHS) of (10) converges to zero as $n$ grows to infinity. For the case of bounded loss functions, it is known that to have convergence, it is sufficient that the *VC-dimension* (to be described shortly) of the class of loss functions is finite [3], [4], [5]. Hence, provided that the class of loss functions has a finite VC-dimension, the decision policy is universally (distribution free) consistent and the convergence has an exponential behavior. For the case of unbounded loss functions, however, universal consistency does not hold, and additional analytical restrictions should be applied to the class of loss functions, $\mathscr{Q}$, and the collection of probability measures, $\mathscr{P}_{\Theta,\mathcal{H}}$.

## 3.1 Growth function

Uniform convergence property is closely dependent on the existence of certain simplifying structures on $\mathcal{Q}$ that restrict the *growth function* of the class of loss functions [4].

Let $L = \inf_{o,g} Q(o, g)$. For every $c > 0$ and every $Q(\cdot, g)$, define a function $Q_c \colon \mathcal{O} \times \mathcal{G} \to \mathbb{R}$ such that

$$Q_c(o, g) = \min\{Q(o, g), c + L\}. \tag{16}$$

Let $\mathcal{Q}_c = \{Q_c(\cdot, g) : Q(\cdot, g) \in \mathcal{Q}\}$ denote the class of truncated loss functions. For every pair $c$ and $g$, let $\mathcal{O}_{c,g}(\alpha) = \{o \in \mathcal{O} : Q_c(o, g) \geq \alpha\}$ denote the *level set* of $Q_c(o, g)$ corresponding to level $\alpha \geq 0$. Let I denote the binary indictor function. Suppose a sequence of observations $o^n = (o_1, o_2, \ldots, o_n)$ is given. Let define

$$\mathcal{Q}_c(o^n) = \left\{(v_{1,\alpha}, \cdots, v_{n,\alpha}) \in \{0,1\}^n : v_{i,\alpha} = \mathrm{I}(o_i \in \mathcal{O}_{c,g}(\alpha)), \forall \alpha \geq L, g \in \mathcal{G}\right\}. \tag{17}$$

Intuitively, $\mathcal{Q}_c(o^n)$ is the set of all distinct vertices on $\{0,1\}^n$ that are spanned by the image of the observation vectors under the characteristic functions of the level sets. Let $N(\mathcal{Q}_c, o^n)$ denote the cardinality of $\mathcal{Q}_c(o^n)$. The growth function of $\mathcal{Q}_c$ is defined

$$G(\mathcal{Q}_c, n) = \max_{o^n \in \mathcal{O}^n} \ln N(\mathcal{Q}_c, o^n). \tag{18}$$

The VC-dimension, named after Vapnik and Chervonenkis, of $\mathcal{Q}_c$ is defined

$$d_c = \max\{n \in \mathbb{N} : G(\mathcal{Q}_c, n) = n \ln 2\} \tag{19}$$

and described as follows. For a sequence of observations $o^n$, let $B$ denote the set of its elements. $B$ is said *shattered* by $\mathcal{Q}_c$, if for every subset $E$ of its elements, there exists $\alpha$ and $g$ such that for every $o \in E$, $Q_c(o, g) \geq \alpha$, and for every element $o \in B - E$, $Q_c(o, g) < \alpha$. The VC-dimension of $\mathcal{Q}_c$ is the cardinality of the largest set $B$ that can be shattered by $\mathcal{Q}_c$. If $\mathcal{Q}_c$ has a finite VC-dimension $d_c$, then it is known that [4, Thm. 4.3]

$$\begin{aligned} G(\mathcal{Q}_c, n) &= n \ln 2, \quad \text{if } n \leq d_c, \\ G(\mathcal{Q}_c, n) &\leq d_c \ln \frac{en}{d_c}, \quad \text{otherwise.} \end{aligned} \tag{20}$$

If $d_c$ is uniformly bounded over $c$, then $d = \lim_{c \to \infty} d_c$ and

$$G(\mathcal{Q}, n) = \limsup_{c \to \infty} G(\mathcal{Q}_c, n)$$

denote the VC-dimension and the growth function of $\mathcal{Q}$, respectively.

**Lemma 3.2.** *For an action $g$, let $J_{c,\theta}(h, g)$ and $J_c^{(n)}(g)$ denote the risk value and empirical risk value that are obtained using the truncated loss functions,*

$Q_c(\cdot, g)$, respectively. Then,

$$P_\theta(\omega \in \Omega : \sup_{g \in \mathcal{G}} |J_{c,\theta}(h,g) - J_c^{(n)}(g)| \geq \varepsilon/2)$$
$$\leq 4\exp\left\{\left(\frac{G(\mathcal{Q}_c, 2n)}{n} - \frac{\varepsilon_*^2}{4c^2}\right)n\right\} \quad (21)$$

for every $n \geq \frac{8c^2}{\varepsilon^2}$ where $\varepsilon_* = \varepsilon - \frac{2}{n}$.

*Proof.* The proof follows by [4, Theorem 5.1] where the lower bound on $n$ is imposed by a well known result on symmetrization [6, Lemma 11.5]. □

Lemma 3.2 provides a sufficient condition for uniform convergence for the bounded class of truncated loss functions $\mathcal{Q}_c$. It is seen that to have uniform convergence, it suffices that $\limsup_{n\to\infty} \frac{G(\mathcal{Q}_c, 2n)}{n} = 0$. By (20), if the VC-dimension of $\mathcal{Q}_c$ is finite, then (21) converges to zero and uniform convergence occurs. A generalization of VC-dimension is also known to be both necessary and sufficient for uniform convergence [5]. Note that the asymptotic uniform convergence behavior for $Q_c$ is $O(e^{-(\gamma_c n)^{q_c}})$ where the *decay exponent* is $q_c = 1$ and the *decay constant* is $\gamma_c = \frac{\varepsilon^2}{4c^2}$.

## 3.2 Dominating function

To generalize Lemma 3.2 to unbounded loss functions, let $M \colon \mathcal{O} \to \mathbb{R}^+$ be a measurable function such that for all $o \in \mathcal{O}$ and $g \in \mathcal{G}$, $Q(o,g) - L \leq M(o)$. Such a function is called a *dominating function* for $\mathcal{Q}$. To have exponential convergence in (10), we need to apply some restricting conditions on $M$.

**Definition 3.1 (Orlicz expectation).** *Let $\psi \colon \mathbb{R}^+ \to \mathbb{R}^+$ be a measurable, nondecreasing function. For an underlying probability measure P, the quantity*

$$\|M\|_\psi \triangleq \inf\left\{c > 0 : \int \psi(|M|/c)dP \leq 1\right\}. \quad (22)$$

*is called the* Orlicz expectation *of $M$. Moreover, if $\psi$ is convex, then $\|M\|_\psi$ is called the* Orlicz norm *of $M$.*[1]

To emphasize the tail behavior of density functions, we consider a special form of functions $\psi_p : \mathbb{R} \to \mathbb{R}^+$, $p > 0$, such that $\psi_p(x) = e^{|x|^p} - 1$. The following result, which is adopted from [7] and generalized for any $p > 0$, motivates the usage of such functions. We call $p$ and $\|M\|_{\psi_p}$ the *exponential expectation order* and the *exponential Orlicz expectation of $M$ of order $p$*, respectively.

**Lemma 3.3.** *For any probability measure $P \in \mathscr{P}_{\Theta, \mathcal{H}}$ and any $p > 0$, the following are equivalent:*

1. $\|M\|_{\psi_p} < \infty$.

---
[1]One special class of Orlicz norms is the class of $L_p$-norms.

2. There exists constants $0 < R, S < \infty$ such that

$$P(|M| > c) \leq Re^{-Sc^p}, \text{ for all } c > 0. \quad (23)$$

Moreover, if either condition holds, then $R = 2$ and $S = \|M\|_{\psi_p}^{-p}$ satisfies (23).

*Proof.* Suppose the first statement holds. Then, by Markov's inequality, we have

$$\begin{aligned}
P(|M| > c) &\leq P\Big(\psi_p(|M|/\|M\|_{\psi_p}) \geq \psi_p(c/\|M\|_{\psi_p})\Big) \\
&\leq \min\left\{\frac{1}{e^{\|M\|_{\psi_p}^{-p} c^p} - 1}, 1\right\} \leq 2e^{-\|M\|_{\psi_p}^{-p} c^p}.
\end{aligned}$$

Now, suppose the second statement holds. Using Fubini's theorem [8] to exchange the order of integration, we obtain

$$\begin{aligned}
\int (e^{c^{-p} M^p} - 1) dP &= \int \int_0^{c^{-p} M^p} e^x dx dP \\
&= \int_0^\infty P(M > cx^{1/p}) e^x dx \\
&\leq \int_0^\infty Re^{-(Sc^p - 1)x} dx = \frac{R}{Sc^p - 1}
\end{aligned}$$

that means for $c \geq \left(\frac{R+1}{S}\right)^{1/p}$ the integration in the LHS is no larger than one. This indicates that $\|M\|_{\psi_p} \leq \left(\frac{R+1}{S}\right)^{1/p} < \infty$. □

In other words, Lemma 3.3 means that to have the probability of the level sets of the dominating function converge to zero, exponentially, it is necessary and sufficient that $\|M\|_{\psi_p} < \infty$ for some $p > 0$. Intuitively, Lemma 3.3 suggests that to obtain an exponential upper bound on (10), it suffices to have $\|M\|_{\psi_p}$ uniformly bounded for some $p > 0$.

### 3.3 Sufficient condition

Using Lemmas 3.2 and 3.3, we now state the key result of this work.

**Theorem 3.1.** *For a given class of loss functions $\mathscr{Q}$ and the collection of probability measures $\mathscr{P}_{\Theta,\mathcal{H}}$, suppose there exists a dominating function $M$, $p > 0$, and $\rho < \infty$ such that $\sup_{P \in \mathscr{P}_{\Theta,\mathcal{H}}} \|M\|_{\psi_p} = \rho$. Then,*

1. *The following inequality holds true*

$$\sup_{\theta \in \Theta} \sup_{h \in \mathcal{H}} P_\theta(\omega \in \Omega : \sup_{g \in \mathcal{G}} |J_\theta(h, g) - J^{(n)}(g)| \geq \varepsilon/2)$$
$$\leq \Big(2n + 4\exp\{G(\mathscr{Q}_{c(n)}, 2n)\}\Big) \exp\left\{-\frac{(1 - \frac{4}{\varepsilon n})^2 \varepsilon^{2q}}{16^q \rho^{2q}} n^q\right\} \quad (24)$$

for $n \geq \frac{16\rho^2}{\varepsilon^2} \max\left\{2^{\frac{2+p}{p}}, \left(\ln \frac{16\rho^{2p}}{\varepsilon p}\right)^{\frac{2+p}{p}}\right\}$ where $c(n) = 16^{-\frac{1}{2+p}} \rho^{\frac{p}{2+p}} \varepsilon^{\frac{2}{2+p}} n^{\frac{1}{2+p}}$ and $q = \frac{p}{2+p}$.

2. If $\limsup_{n \to \infty} \frac{G(\mathcal{Q}_{c(n)}, 2n)}{n^q} = 0$, then the empirical risk minimization approach is consistent.

*Proof.* Without the loss of generality, we may assume that $L = 0$, otherwise, take $Q = Q - L$. For a given loss function $Q$ and a positive value $c > 0$, let $Q_c$ be defined as (16) and $Q^c \triangleq Q - Q_c$. Moreover, let $J_{c,\theta}(h,g) = \int Q_c(o,g) dP_{\theta,h}$ and $J_\theta^c(h,g) = \int Q^c(o,g) dP_{\theta,h}$. Let $I_c^{(n)}(g)$ and $I^{c,(n)}(g)$ be defined, similarly. One can verify that for every $c > 0$, we have

$$P_\theta(\omega \in \Omega : \sup_{g \in \mathcal{G}}|J_\theta(h,g) - J^{(n)}(g)| \geq \varepsilon/2)$$
$$\leq P_\theta(\omega \in \Omega : \sup_{g \in \mathcal{G}}|J_{c,\theta}(h,g) - J_c^{(n)}(g)| \geq \varepsilon/4)$$
$$+ P_\theta(\omega \in \Omega : \sup_{g \in \mathcal{G}}|J_\theta^c(h,g) - J^{c,(n)}(g)| \geq \varepsilon/4). \quad (25)$$

For the dominating function $M$ let define functions $M_c = \min(M, c)$ and $M^c = M - M_c$. For the second term of (25), we note that

$$P_\theta(\omega \in \Omega : \sup_{g \in \mathcal{G}}|J_\theta^c(h,g) - J^{c,(n)}(g)| \geq \varepsilon/4)$$
$$\leq P_\theta(\omega \in \Omega : \int M^c dP_{\theta,h} \geq \varepsilon/8)$$
$$+ P_\theta(\omega \in \Omega : \int M^c dP^{(n)} \geq \varepsilon/8). \quad (26)$$

It can be verified that

$$P_\theta(\omega \in \Omega : \int M^c dP_{\theta,h} \geq \varepsilon/8) \leq P_\theta(\omega \in \Omega : \int_{M(o)>c} M dP_{\theta,h} \geq \varepsilon/8)$$
$$= P_\theta(\omega \in \Omega : \int P_{\theta,h}(M \geq c+x) dx \geq \varepsilon/8)$$
$$\text{(By Lemma 3.3)} \leq P_\theta(\omega \in \Omega : \int 2e^{-\rho^{-p}(c+x)^p} dx \geq \varepsilon/8)$$
$$\leq P_\theta(\omega \in \Omega : \frac{2\rho^p e^{-\rho^{-p}c^p}}{c^{p-1}p} \geq \varepsilon/8),$$

where by taking $c > c_{\text{th}} \simeq \rho \sqrt[p]{\ln \frac{16\rho^{2p}}{\varepsilon p}}$, we enforce the LHS to be zero. Now, consider the second term of the RHS of (26). We have

$$P_\theta(\omega \in \Omega : \int M^c dP^{(n)} \geq \varepsilon/8) \leq nP_\theta(\omega \in \Omega : M(O(\omega)) \geq c + \varepsilon/8)$$
$$\text{(By Lemma 3.3)} \leq 2n \exp(-\rho^{-p}c^p). \quad (27)$$

Moreover, by applying Lemma 3.2 to the first term of the RHS of (25), we have

$$P_\theta(\omega \in \Omega : \sup_{g \in \mathcal{G}} |J_{c,\theta}(h,g) - J_c^{(n)}(g)| \geq \varepsilon/4) \leq 4 \exp\left\{\left(\frac{G(\mathcal{Q}_c, n)}{n} - \frac{\epsilon_*^2}{16c^2}\right)n\right\} \quad (28)$$

for every $n \geq \frac{32c^2}{\varepsilon^2}$, where $\epsilon_* = \varepsilon - 4/n$. Provided that $c \geq c_{\text{th}}$, the asymptotic behavior of the upper bound is controlled either by (27) or (28). To obtain the tightest bound, we pick $c$ such that $\rho^{-p}c^p \simeq \frac{\epsilon_*^2 n}{16c^2}$. As a result, we obtain

$$c = 16^{-\frac{1}{2+p}} \rho^{\frac{p}{2+p}} \varepsilon^{\frac{2}{2+p}} n^{\frac{1}{2+p}}.$$

Note that for the case of bounded functions as $p$ increases, $c$ goes to $\rho$. For the general case where $p < \infty$, we need to have $c > c_{\text{th}}$ that requires

$$n \geq \frac{16\rho^2}{\varepsilon^2} \left(\ln \frac{16\rho^{2p}}{\varepsilon p}\right)^{\frac{2+p}{p}}.$$

Moreover, substituting this $c$ in condition $n \geq \frac{32c^2}{\varepsilon^2}$, one can verify that for

$$n \geq \frac{16\rho^2}{\varepsilon^2} \max\left\{2^{\frac{2+p}{p}}, \left(\ln \frac{16\rho^{2p}}{\varepsilon p}\right)^{\frac{2+p}{p}}\right\},$$

the following inequality holds true

$$\sup_{\theta \in \Theta} \sup_{h \in \mathcal{H}} P_\theta(\omega \in \Omega : \sup_{g \in \mathcal{G}} |J_\theta(h,g) - J^{(n)}(g)| \geq \varepsilon/2) \leq$$
$$\left(2n + 4 \exp\{G(\mathcal{Q}_c, n)\}\right) \exp\left\{-\frac{(1-\frac{4}{\varepsilon n})^2 \varepsilon^{2q}}{16^q \rho^{2q}} n^q\right\},$$

where $q = \frac{p}{2+p}$.

The proof of the second assertion follows by the fact that the given hypothesis provides a sufficient condition for convergence of the RHS of (24) to zero. This convergence means that the ERM policy is $\varepsilon$-consistent. Because $c = 16^{-\frac{1}{2+p}} \rho^{\frac{p}{2+p}} \varepsilon^{\frac{2}{2+p}} n^{\frac{1}{2+p}}$ is an increasing function with respect to $\varepsilon$, $G(\mathcal{Q}_c, n)$ is non-increasing as $\varepsilon$ decreases. This implies the second assertion. □

Theorem 3.1 embodies key results in assessing uniform convergence and its behavior for decision-making with unbounded loss functions. Equation (24) is an upper bound on the worst case probability of taking an action whose risk is not within $\varepsilon$ accuracy of the minimal risk. Thus, convergence of (24) to zero implies $\varepsilon$-consistency of the decision policy. By second assertion, convergence occurs provided that the collection of truncated loss functions, $\mathcal{Q}_{c(n)}$, satisfies

$$\limsup_{n \to \infty} \frac{G(\mathcal{Q}_{c(n)}, 2n)}{n^q} = 0, \quad (29)$$

where $c(n)$ is described as in the assertion.

By (20), if $\mathscr{Q}$ has a finite VC-dimension, then $G(\mathscr{Q}_{c(n)}, 2n) \leq G(\mathscr{Q}, 2n) = O(\ln n)$ implying that (29) holds true for any $q > 0$. This implies that for an unbounded class of loss functions with finite VC-dimension, the ERM policy is consistent, if there exist a dominating function $M$ and an exponential expectation order $p > 0$ such that $\sup_{P \in \mathscr{P}_{\Theta,\mathcal{H}}} \|M\|_{\psi_p} < \infty$. By Borel-Cantelli Lemma [9], these conditions are also sufficient for strong consistency of the ERM policy.

Upon satisfaction of (29), by (24), it appears that the asymptotic behavior of the convergence is described by $O(e^{-\gamma^q n^q})$, where $\gamma = \frac{\varepsilon^2}{16\rho^2}$ is the decay constant and $q$ is the decay exponent for the case of unbounded loss functions. The decay exponent, $q = \frac{p}{2+p}$, is an increasing function of $p$. We note that $q < 1$ meaning that the decay exponent for unbounded loss functions is strictly smaller than one which is in clear contrast from the case of bounded loss functions as seen in (21). On the other hand, the decay constant $\gamma = \frac{\varepsilon^2}{16\rho^2}$ is very similar to the case of bounded loss functions where the constant $c$ is replaced by $\rho$. Clearly, in both cases, as either $c$ or $\rho$ increase, the decay constant decreases imposing a slower convergence.

Given a parameter $\beta \in [0, 1]$ and a desired level of accuracy $\varepsilon$, *sample complexity* of ERM policy is defined as

$$n(\varepsilon, \beta) = \inf\{n \in \mathbb{N} : p(\boldsymbol{A}_{\mathrm{ERM}}, \varepsilon, n) \leq \beta\}. \tag{30}$$

Sample complexity determines the number of observations needed by ERM decision policy to take actions whose risk is within accuracy $\varepsilon$ of the minimal risk with a *confidence* of at least $1 - \beta$. Assuming that the class of unbounded loss functions, $\mathscr{Q}$, has a finite VC-dimension of size $d$, we obtain $n(\varepsilon, \beta) \leq n_o$, where

$$n_o = \frac{16\rho^2}{\varepsilon^2} \left( \ln \frac{2 + 4e^2}{\beta} + d \ln n_o \right)^{1/q}. \tag{31}$$

Equation (31) is a novel expression that demonstrates the effect of different parameters in determining the sample complexity. It can be seen that the accuracy $\varepsilon$ and the least upper bound on the exponential expectation $\rho$ have more dramatic impact on $n_o$ comparing to the confidence parameter $\beta$. Moreover, as the decay exponent, $q$, increases, the sample complexity decreases. Due to dependency of both $q$ and $\rho$ to the tail behavior of the density functions, it is expected that densities with heavier tail result in smaller $q$ and larger $\rho$, hence, larger sample complexity.

## 4 Applications

As mentioned earlier, we adopted a decision-making framework since it is a generic representation for different applications. To further clarify how the derived results can be used in practice, we now use this model in some example applications.

## 4.1 Method of moments

Parameter estimation through *method of moments* is a realization of *substitution estimators* that is described as follows [10, Ch. II]. Let $\mathcal{X}$ be a subset of a Euclidean space that denotes the observation space. Let $(\Omega, \mathcal{F}, P)$ be a probability space. Let $\mathcal{H} \subset \mathbb{R}^l$ be a compact, convex subset for some integer $l > 0$ such that for every $h \in \mathcal{H}$, $P_h$ is a probability distribution of a random variable $X_h \colon \Omega \to \mathcal{X}$ whose distribution is induced by $P$. The collection of such induced probability distributions is denoted by $\mathscr{P}_\mathcal{H}$. It is assumed that there exists a measurable function $\phi \colon \mathcal{X} \to \mathbb{R}^l$ such that for an underlying probability measure $P_h \in \mathscr{P}_\mathcal{H}$, one obtains $h$ through a multidimensional integration as follows

$$h = \int \phi(x) dP_h. \tag{32}$$

For example, suppose $\mathcal{X} = \mathbb{R}^l$ and function $\phi$ is described as

$$\phi(x) = \begin{bmatrix} |x_1|^m \\ \vdots \\ |x_l|^m \end{bmatrix} \text{ for some } m \in \mathbb{N},$$

which implies that parameter $h$ is the $m$-order moment of the underlying distribution. In practice, since a sequence of observations $x^n = (x_1, x_2, \ldots, x_n)$ is given instead of $P_h$, an estimate of $h$ is obtained by substituting an empirical measure $\hat{P}^{(n)}$ in (32) that generates an estimate

$$\hat{h}^{(n)} = \int \phi(x) d\hat{P}^{(n)} = \frac{1}{n} \sum_{i=1}^{n} \phi(x_i). \tag{33}$$

It can be seen that $\hat{h}^{(n)}$ is not necessarily in $\mathcal{H}$ for all observed sequences. In such incidences, the estimator picks a boundary point $\hat{h}^n \triangleq \arg\min_{h \in \mathcal{H}} \|h - \hat{h}^n\|^2$ as the estimate. For a $\varepsilon > 0$, the estimator is said to be $\varepsilon$-consistent if

$$\limsup_{n \to \infty} \sup_{h \in \mathcal{H}} P_h(\|\hat{h}^{(n)} - h\|^2 > \varepsilon) = 0. \tag{34}$$

Moreover, if

$$\sup_{h \in \mathcal{H}} P_h(\limsup_{n \to \infty} \|\hat{h}^{(n)} - h\|^2 > \varepsilon) = 0, \tag{35}$$

the estimator is said to be strongly $\varepsilon$-consistent. In either case, if the estimator is (strongly) $\varepsilon$-consistent for every $\varepsilon > 0$, it is said to be (strongly) consistent. Sufficient conditions for consistency of these estimators have been previously addressed in the literature [10, Ch. II.4]. Here, we demonstrate how similar results can be easily obtained from the results of this work.

To articulate this problem into the generic model that is used in this work, let $\mathcal{G} = \mathcal{H}$ denote the action space, and let $Q(x, g) = \|\phi(x) - g\|^2$ describe the loss function. The risk (error) of an action (estimate) $g$ when the underlying parameter is $h$ is

$$J(h, g) = \int \|\phi(x) - g\|^2 dP_h. \tag{36}$$

Let $(g_k)$ be any sequence in $\mathcal{G}$ such that $g_k \to g$. By Fatou's Lemma [8],

$$J(h, g) \leq \liminf_{k \to \infty} J(h, g_k).$$

This means that $J(h, g)$ is lower semi-continuous; hence, Definition 2.1 is applicable to the method of moments. Since $\mathcal{G}$ is convex and $Q(x, g)$ is strictly convex over $\mathcal{G}$, the solution of empirical risk minimization, (9), is unique and equal to the solution obtained from the method of moments. Thus, the notion of consistency defined in (35) is equivalent to the one defined in Definition 2.2. As a result, the analytical arguments shown in this work can be used to investigate consistency and exponential convergence behavior of the method of moments. Since the VC-dimension of the class of loss functions is upper bounded by $l+1$, to have an exponential uniform convergence, it suffices to find a dominating function whose exponential Orlicz expectation (for some exponent $p > 0$) is uniformly bounded over $P_{\mathcal{H}}$. Thus, if there exist constants $a, b > 0$ such that for every $P_h \in P_{\mathcal{H}}$, the density function $f_h(x) = O(e^{-a\|\phi(x)\|^b})$, then the dominating function $M(x) = 2\|\phi(x)\|^2 + 2\sup_{g \in \mathcal{G}} \|g\|^2$ satisfies the condition of Theorem 3.1 for every $p < b/2$.

This basically means that to verify exponential convergence of the method of moments, one can simply check the existence of constants $a, b$ as described above. Furthermore, for a convenient exponential Orlicz expectation order $p$, by estimating the value of $\rho$, one can investigate decay rate and sample complexity of the method of moments. Similarly, using the methodology shown here, one can articulate *M-estimators* [10] into the decision-making framework, and use the results of this work to obtain important insights about consistency and convergence behavior, hence, strong consistency of such estimators.

## 4.2 System identification

Example 2.1 presents a basic system identification problem where the risk function is the mean square error. Suppose the probabilistic model $\mathscr{P}_\Theta$ is a collection of probability measures such that each measure $P_\theta \in \mathscr{P}_\Theta$ has a density function $f_\theta(x, v) = k(\theta) e^{-\frac{1}{2}\langle x, \theta^{-1} x \rangle - \frac{1}{2}|v|^2}$, where $\theta$ is a positive definite matrix of size $m + 1$ whose eigenvalues are in the interval $[a, b]$ for $0 < a < b$ and $k(\theta) = (2^{m+1} \pi^{m+1} \det \theta)^{-1/2}$. For a pair of underlying $\theta$ and $h$, the density function of the induced probability measure, $P_{\theta,h}$, on the space of observations, $\mathcal{X} \times \mathcal{Y}$, is described by

$$f_{\theta,h}(x, y) = k(\theta) e^{-\frac{1}{2}\langle x, \theta^{-1} x \rangle - \frac{1}{2}|y - \langle h, x \rangle|^2}.$$

Since $\mathcal{H}$ is compact the collection of probability measures $\mathscr{P}_{\Theta, \mathcal{H}}$ is a tight collection.

Recall that for a given observation pair $(x,y)$, the cost of taking action $g \in \mathcal{G}$ is $Q((x,y),g) = |y - \langle g, x \rangle|^2$. Since $\mathcal{G}$ is compact, $\sup_{g \in \mathcal{G}} \|g\| = \Lambda$ for some $\Lambda < \infty$. Hence, one can simply define the dominating function as $M(x,y) = 2|y|^2 + 2\Lambda^2 \|x\|^2$. Thus, for any exponential expectation order $p < 1$, it can be verified that $\sup_{P \in \mathscr{P}_{\Theta,\mathcal{H}}} \|M\|_{\psi_p} = \rho$ for some $\rho < \infty$. Since the VC-dimension of the class of loss functions is $d = m+1$ [4, Ch. 5], the second assertion of Theorem 3.1 holds true. This means that for this setup the empirical minimum square error system identification is consistent. Furthermore, by estimating the value of $\rho$, the decay rate of convergence can be estimated.

### 4.3 Maximum likelihood estimation

Example 2.2 articulates a generic maximum likelihood estimation problem in the framework of decision-making. Let $\mathcal{X}$ be the positive cone in $\mathbb{R}^m$, and let $\mathcal{H} = [a,b]^m$ where $0 < a < b < \infty$. For probability measure $P_h \in \mathscr{P}_{\mathcal{H}}$, let

$$f_h(x) = k(h) e^{-\langle h, x \rangle}$$

describe its density function where $k(h) = \prod_{i=1}^n h_i$. As a result, the cost of taking action $g$ for observation $x$ is $Q(x,g) = -\ln k(g) + \langle g, x \rangle$. Taking the dominating function as

$$M(x) = m(|\ln a| + |\ln b|) + \sqrt{m} b \|x\|,$$

one can verify that $\sup_{P \in \mathscr{P}_{\mathcal{H}}} \|M\|_{\psi_p} < \infty$ for every exponential expectation order $p < 1$. Since the VC-dimension of the class of loss functions is $d = m$, by Theorem 3.1, the empirical maximum likelihood estimation is consistent for this setup. With some additional effort to estimate $\rho$, one would be able to obtain important insights regarding the convergence behavior of the maximum likelihood estimation.

## 5 Conclusion

The problem of decision-making with side information using unbounded loss functions was considered. A decision-making model was used as a generic system-theoretic representation for a broader range of applications in machine learning, signal processing, and communications. Sufficient conditions for consistency of the ERM decision policy with exponential convergence behavior were derived. These conditions are also sufficient for strong consistency of the ERM decision policy with unbounded loss functions. It was shown that including a condition on the growth rate of the class of loss functions, it suffices to have a dominating function whose exponential Orlicz expectation is uniformly bounded over the probabilistic model. The results verify that the decay constant is similar to the decay constant known for the case of bounded loss functions, but the decay exponent is strictly smaller than one.